\pdfoutput=1

\documentclass[11pt]{article}
\usepackage{tikz}
\usepackage[preprint]{acl}

\usepackage{times}
\usepackage{latexsym}

\usepackage[T1]{fontenc}

\usepackage[utf8]{inputenc}
\usepackage{CJKutf8}
\usepackage{microtype}

\usepackage{inconsolata}

\usepackage{graphicx}
\usepackage{subfigure}
\usepackage{booktabs}
\usepackage{makecell}
\usepackage{multirow}
\usepackage[utf8]{inputenc}
\usepackage[T1]{fontenc}
\usepackage{CJKutf8}
\usepackage{comment}
\usepackage[most]{tcolorbox}
\tcbset{
    userstyle/.style={
        enhanced,
        colback=white,
        colframe=black,
        colbacktitle=gray!20,
        coltitle=black,
        rounded corners,
        sharp corners=north,
        boxrule=0.5pt,
        drop shadow=black!50!white,
        attach boxed title to top left={
            xshift=-2mm,
            yshift=-2mm
        },
        boxed title style={
            rounded corners,
            size=small,
            colback=gray!20
        }
    },
    replystyleg/.style={
        enhanced,
        colback=green!15,
        colframe=black,
        colbacktitle=green!30,
        coltitle=black,
        boxrule=0.5pt,
        drop shadow=black!50!white,
        rounded corners,
        sharp corners=north,
        attach boxed title to top right={
            xshift=-2mm,
            yshift=-2mm
        },
        boxed title style={
            rounded corners,
            size=small,
            colback=green!40
        }
    },
    replystyler/.style={
        enhanced,
        colback=red!15,
        colframe=black,
        colbacktitle=red!40,
        coltitle=black,
        boxrule=0.5pt,
        drop shadow=black!50!white,
        rounded corners,
        sharp corners=north,
        attach boxed title to top right={
            xshift=-2mm,
            yshift=-2mm
        },
        boxed title style={
            rounded corners,
            size=small,
            colback=red!40
        }
    }
}
\newtcolorbox{userquery}[1][]{
    userstyle,
    title=Prompt,
    #1
}

\title{C-FAITH: A Chinese Fine-Grained Benchmark for \\
Automated Hallucination Evaluation}

\author{Xu Zhang, ~~Zhifei Liu, ~~Jiahao Wang,~~Huixuan Zhang,\\ 
    \bf ~~Fan Xu,~~Junzhe Zhang \and  Xiaojun Wan \\
        Wangxuan Institute of Computer Technology, Peking University
        }

\begin{document}
\maketitle
\begin{abstract}
Despite the rapid advancement of large language models, they remain highly susceptible to generating hallucinations, which significantly hinders their widespread application.
Hallucination research requires dynamic and fine-grained evaluation.
However, most existing hallucination benchmarks (especially in Chinese language) rely on human annotations, making automatical and cost-effective hallucination evaluation challenging.
To address this, we introduce HaluAgent, an agentic framework that automatically constructs fine-grained QA dataset based on some knowledge documents.
Our experiments demonstrate that the manually designed rules and prompt optimization can improve the quality of generated data.
Using HaluAgent, we construct C-FAITH, a Chinese QA hallucination benchmark created from 1,399 knowledge documents obtained from web scraping, totaling 60,702 entries.
We comprehensively evaluate 16 mainstream LLMs with our proposed C-FAITH, providing detailed experimental results and analysis.
\end{abstract}

\section{Introduction}
Despite significant advances made by large language models (LLMs) \citep{llama3, gpt4} in natural language generation, hallucination continues to undermine their reliability and safety \citep{xu2024hallucinationinevitableinnatelimitation, Huang_2024}.
The issue of hallucination makes the deployment of LLMs potentially risky in real-world applications \citep{bang-etal-2023-multitask, 10.1145/3571730}.
To understand what types of content and to what extent LLMs tend to hallucinate, much attention has been paid to constructing high-quality datasets for hallucination evaluation \citep{chinesefacteval, chinesesimpleqa}.
Since the potential hallucinations of LLMs exist in various domains, the size and scalability of datasets are crucial for the oversight of LLM hallucinations. 

However, constructing and scaling-up hallucination evaluation datasets face significant challenges   \citep{cao2024autohallautomatedhallucinationdataset, liu2024phdchatgptpromptedvisualhallucination, anahv2}.
As existing hallucination benchmarks \citep{halu_eval, chen2024factchd} often rely on human annotations to construct high-quality datasets, one primary challenge is the prohibitively high costs of human annotation required for hallucination benchmark construction.
Since manually constructing hallucination benchmarks is time-consuming and expensive, there is a need to develop automatic approaches to construct hallucination evaluation datasets at scale.
To automate the construction of hallucination evaluation dataset, we propose \textbf{HaluAgent}, an agentic framework for automatic dataset generation.

\begin{figure}[t]
    \centering
    \includegraphics[width=1.0\linewidth]{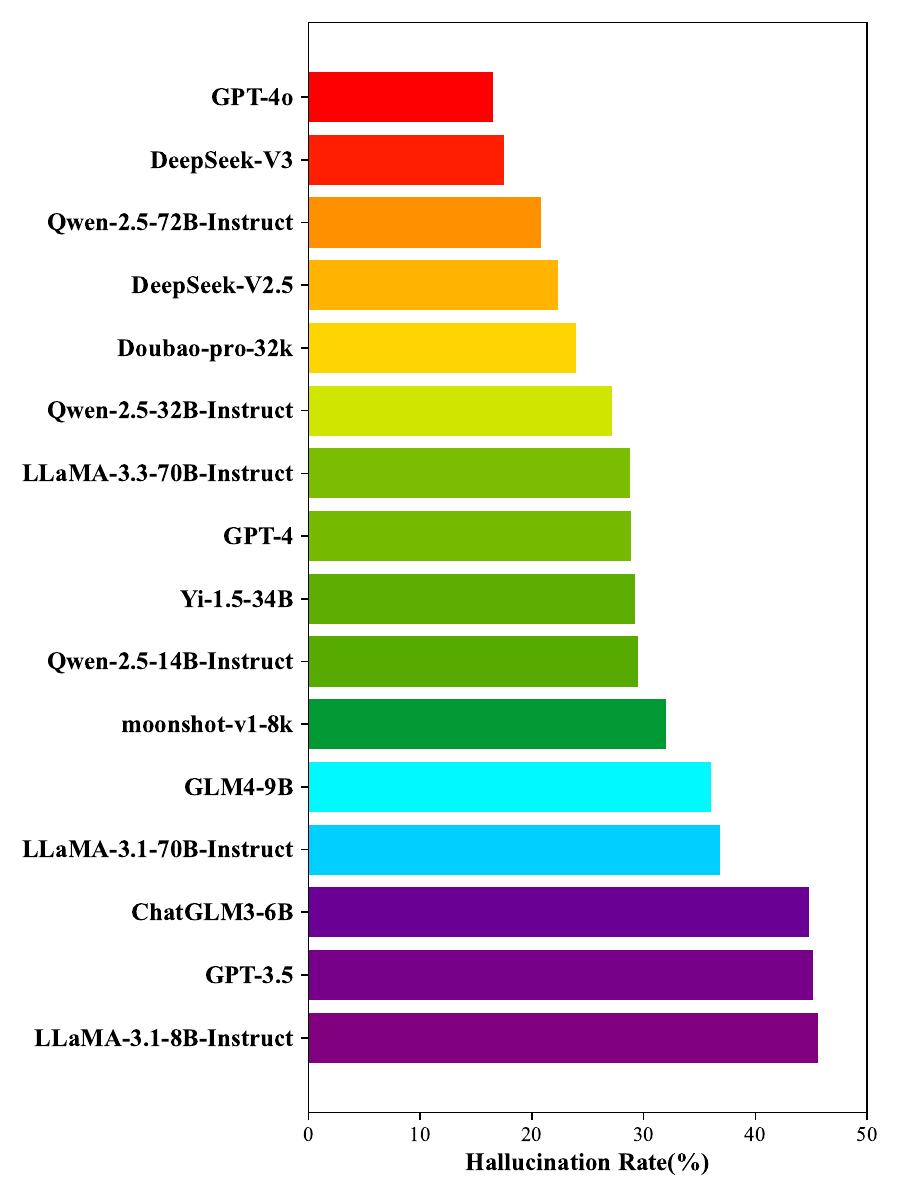}
    \caption{The total hallucination rates of 16 tested LLMs on C-FAITH.}
    \vspace{-5mm}
    \label{fig:eval_results}
\end{figure}

\begin{figure*}[t]
\centering
    \includegraphics[width=0.98\linewidth]{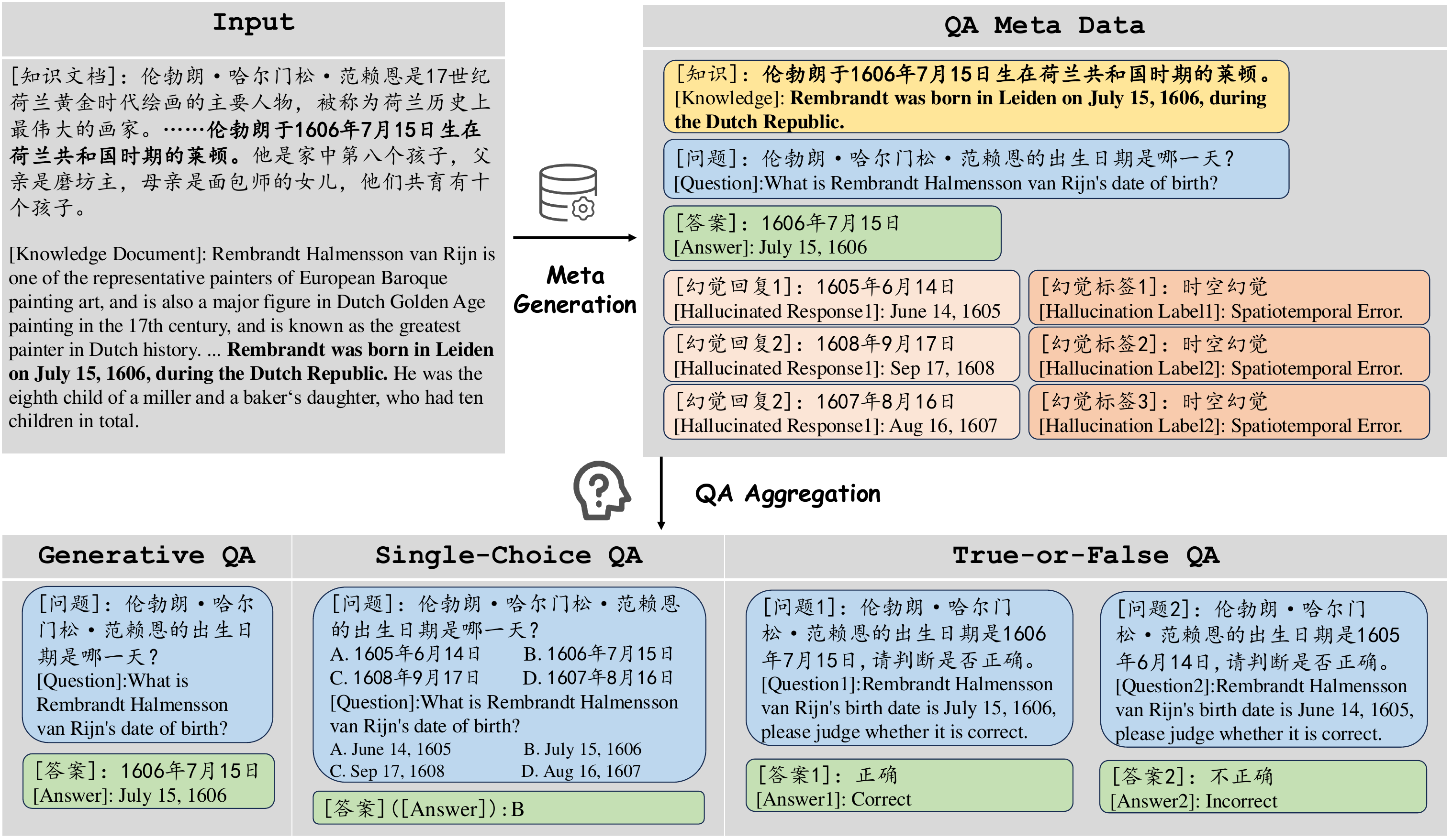}
    \caption{An example of the created QA data. HaluAgent first generates meta data containing question, correct answer, hallucinated responses and hallucination labels. Then, the QA meta data is aggregated into three different formats for hallucination evaluation. We provide both the Chinese QA data and the English translation in the figure.}
    \vspace{-5mm}
\label{fig:halu_intro}
\end{figure*}

We highlight key differences between existing hallucination benchmark construction and our HaluAgent method below:
\begin{itemize}
    \item \textbf{Automatic data constuction:} Unlike some of the previous works that rely on manual annotation for data construction,  we use a multi-agent system, HaluAgent. Built on Qwen model \citep{qwen2}, HaluAgent automatically generates and verifies QA data for hallucination evaluation.
    \item \textbf{Fine-grained error types:} Previous studies classify QA data based on question patterns \citep{chen2024factchd} or topic \citep{anah}. 
    Our study induces different types of hallucination given fine-grained hallucination error types. 
    By generating hallucination labels for the QA data, HaluAgent enables more targeted identification of LLM hallucination.
    \item \textbf{Prompt optimization:} We leverage prompt optimization techniques \citep{yang2024largelanguagemodelsoptimizers} to refine prompts for data generation.
    This improves the quality of data generation over methods that use simple few-shot prompting.
    \item \textbf{Multiple QA formats:} Previous studies often generate a single form of QA pairs, while HaluAgent supports multiple QA formats.
    Different forms of QA data help identify vulnerabilities of LLMs.
\end{itemize}

\begin{table*}
\centering
\begin{tabular}{p{4cm}p{11cm}}
    \toprule
    \small 
    Factual Fabrication(\textbf{FactFab}) & 
    \small
    The LLM fabricates concepts that do not exist or make up facts that do not exist in the real world. \\
    \hline
    \small
    Attribute Error(\textbf{AttrErr}) & 
    \small
    The LLM generates incorrect content when describing object in the real world, such as the composition and function of an object. \\
    \hline
    \small
    Entity Error(\textbf{EntErr}) & 
    \small
    The LLM generates text that contains false entities that contradict the world knowledge, such as people, event names, movies, books. \\
    \hline
    \small
    Relation Error(\textbf{RelErr}) & 
    \small
    The LLM generates text containing false relationships between entities such as quantity, space, time, etc. \\
    \hline
    \small
    Spatiotemporal Error(\textbf{SpaErr}) & 
    \small
    The LLM generates incorrect information about the time and space of an event. \\
    \hline 
    \small
    Reference Error(\textbf{RefErr}) & 
    \small
    The LLM makes up references and links that do not exist to make the generation more reliable. \\
    \bottomrule
    \end{tabular}
    \caption{Classification of hallucinations in LLMs.}
    \label{tab:classification}
\end{table*}

Given knowledge documents as input, HaluAgent supports the generation of three different QA data formats, including generative QA, single-choice QA and true-or-false QA. 
Figure \ref{fig:halu_intro} shows an example of generated data with a piece of knowledge extracted from the input document.
Our HaluAgent method works by first generating questions and the corresponding correct answers.
Then, we generate hallucinated responses for the questions, which are inconsistent with the background knowledge as well as the correct answers.
The hallucinated response represents a potentially hallucinated generation.
Finally, we produce hallucination labels for each hallucinated response to indicate the hallucination type induced by the question.
By combining these elements, we obtain the QA meta-dataset.
Three different QA formats can then be derived from this meta-dataset.

A verification module is proposed to check the correctness of the generated answers, the hallucinated responses, and the hallucination labels.
During the large-scale data generation process, validated QA data are retained as the final hallucination evaluation dataset.
Any data flagged as uncertain by the verification module is filtered out to ensure quality and accuracy.
Moreover, to increase the validation rate of the generated data, we perform prompt optimization before large-scale generation.

As most existing hallucination benchmarks focus on English corpora, the number of QA datasets for Chinese hallucination evaluation is relatively limited.
Therefore, we construct a hallucination benchmark to facilitate hallucination evaluation in the Chinese language in this paper.
We collect 1,399 Chinese knowledge documents from multiple domains to construct a Chinese hallucination benchmark.
Building on HaluAgent, we introduce \textbf{C-FAITH}, a benchmark comprising 16,713 generative QA items, 10,563 single-choice QA items and 33,426 true-or-false QA items in general.
We evaluate 16 mainstream LLMs with C-FAITH, including both open-source and black-box models.
Figure \ref{fig:eval_results} summarizes the total halucination rates of 16 tested LLMs on our proposed C-FAITH.
Moreover, we analyze the hallucination rates of LLMs when faced with different types of questions.
The experimental results indicate that LLMs are most prone to hallucination involving entity errors and spatiotemporal errors.

In summary, our contributions can be listed as follows\footnote{\href{https://github.com/pkulcwmzx/C-FAITH}{https://github.com/pkulcwmzx/C-FAITH}}:
\begin{itemize}
    \item We propose \textbf{HaluAgent}, an automated framework for generating hallucination evaluation datasets with fine-grained error types in different formats.
    \item We introduce \textbf{C-FAITH}, a new Chinese hallucination evaluation dataset with fine-grained error types designed for the systematic assessment of hallucination generated by LLMs.
    \item We evaluate and analyze the hallucination risks of 16 mainstream LLMs with \textbf{C-FAITH}, providing detailed experimental results and analysis.
\end{itemize}

\section{Related Work}
Hallucination benchmarks construct challenging queries in single or multiple tasks to assess hallucination rate in LLM responses. 
These benchmarks cover a wide range of topics and tasks \citep{elaraby2023halo, pal2023med, luo2024halludial, liu2024exploring}.
There are benchmarks curated with semi-automated approaches for data generation \citep{chen2024factchd, chen2024diahalu, anah, mishra2024fine} to offer better expandability compared with datasets that rely solely on manual annotations \citep{chinesefacteval, cheng2023evaluating}.
UHGEval \citep{uhgeval} automates the construction of hallucination benchmarks for text continuation task.
C-FAITH provides an automated, fine-grained and scalable hallucination benchmark for QA task.

Another line of work involves training a hallucination detector to evaluate the hallucination level of LLM generation \citep{muhlgay2023generating, sriramananllm, anahv2, hallumeasure}.
Some early studies \citep{wang-etal-2020-asking, feqa, liu2021token, dziri2022faithdial, gupta-etal-2022-dialfact, laban-etal-2022-summac, varshney2023stitchtimesavesnine, yang-etal-2023-new-benchmark} focus on distinguishing whether the LLM output contains hallucinated content.
Recent researches \citep{mishra2024finegrainedhallucinationdetectionediting, anah} further detect hallucinations in a more fine-grained and meticulous way.

\begin{figure*}[t]
    \centering
    \includegraphics[width=0.9\linewidth]{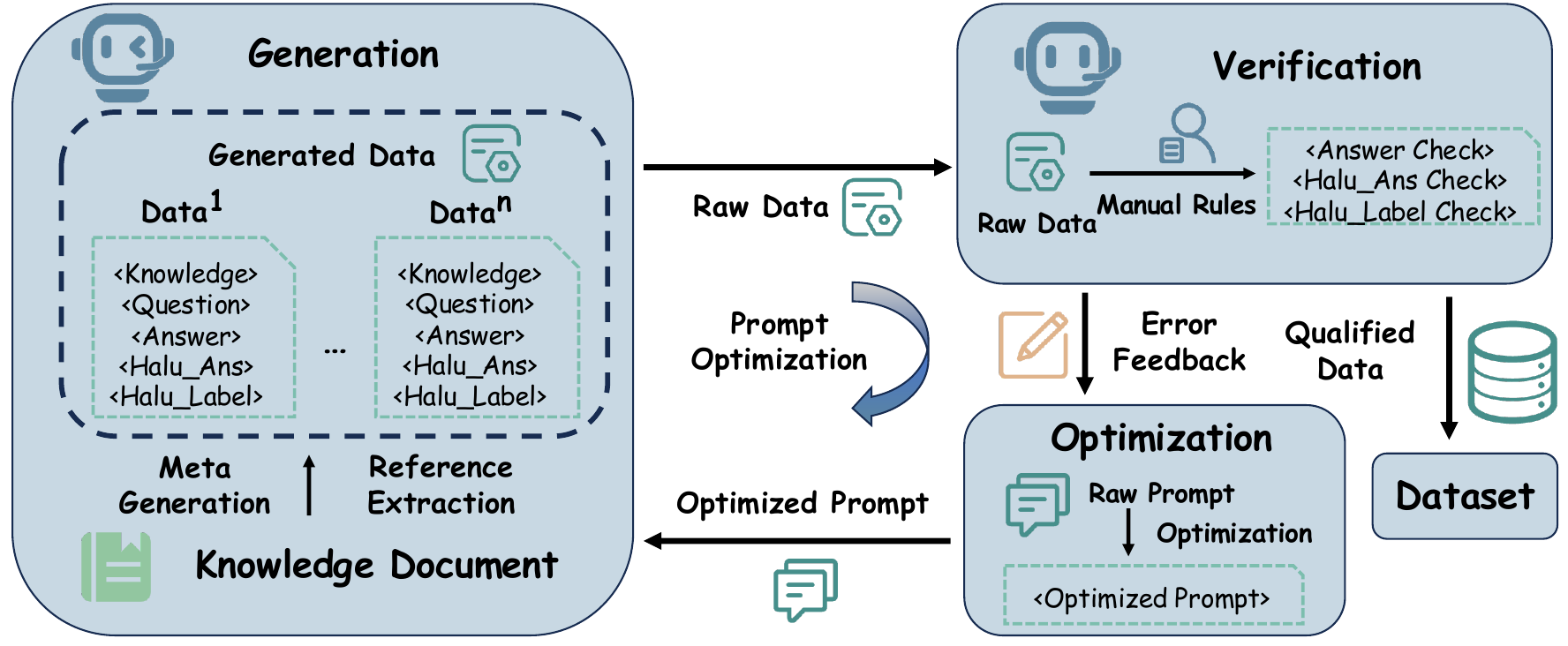}
    \caption{An illustration of our proposed HaluAgent framework. HaluAgent consists of three modules, including the generation module, the verification module and the optimization module. With manually designed rules, HaluAgent first conducts prompt optimization based on error feedbacks from the verification module. Next, HaluAgent takes knowledge documents as input to generate fine-grained QA data for hallucination evaluation.}
    \label{fig:halu_agent}
    \vspace{-5mm}
\end{figure*}

\section{Method}
\label{sec:haluagent}
In this section we present our HaluAgent method.
We use an agentic framework for data generation inspired by recent advances in role-playing \citep{park2023generativeagentsinteractivesimulacra} and prompt optimization \citep{yang2024largelanguagemodelsoptimizers}.
HaluAgent incorporates automated agents based on an open-source model Qwen-2-72B-Instruct\footnote{Other LLMs can be used to construct the evaluation data as well.} \citep{qwen2} to extract QA pairs from knowledge documents, generate hallucinated responses, and classify these hallucinated responses.
Then, HaluAgent validates the correctness of the generated answers, hallucinated responses, and hallucination labels based on predefined manual rules.
With the generated meta dataset, we aggregate it into different formats of QA data to construct the evaluation dataset.

To improve the quality of the generation data, we perform prompt optimization based on error feedbacks to refine the generation prompt before large-scale data generation.
We then aggregate the generated data into multiple QA formats.
Below, we discuss the key components of our HaluAgent.

\subsection{Automatic Generation with Agentic Framework}
Several recent works on hallucination benchmarks leverage LLMs to help create evaluation data.
However, such methods \citep{halu_eval} require manual annotation to modify and validate LLM-generated outputs.
As illustrated in Figure \ref{fig:halu_agent}, our approach decomposes data construction into generation and verification modules.
HaluAgent automatically generates QA data and verifies the correctness of the generated data with predefined strict manual rules.

By prompting the Qwen-2-72B-Instruct model, we build multiple data-generation agents and verification agents to generate and check the data.
For each agent, the prompt structure begins with an initial instruction that specifies the task, followed by a set of rules outlining the task requirements.
It also includes a selection of example inputs accompanied by their manually constructed outputs.
Finally, the prompt ends with the target input for which the agent needs to generate a target output or check the given inputs.
By providing this comprehensive prompt, we aim to teach the model to generate data according to the requirements and verify it based on the input rules.

The generation process is divided into three parts.
Firstly, HaluAgent extracts knowledge from the knowledge document and generates the question and its correct answer.
Then, HaluAgent generates hallucinated responses using the question and correct answer as input.
Hallucinated responses are responses to the question that are inconsistent with the provided sSfacts.
Finally, HaluAgent takes the hallucinated responses and the correct answer as input to generate the hallucination label for each hallucinated response.
We currently distinguish six types of LLM hallucination (Table \ref{tab:classification}).
Different types of hallucinations are exemplified in Table \ref{tab:halu_type} in the appendix.
The generated hallucination label specifies the type of hallucination in the hallucinated responses.

To check the correctness of the generated QA data, we introduce the verification module within HaluAgent.
Similar to the generation process, the verification process consists of three parts, namely correctness check, hallucination check and label check.
Firstly, HaluAgent checks whether the correct answer is a valid response to the question based on the extracted knowledge.
Then, HaluAgent checks whether the generated hallucinated responses are responses that contain hallucinated content.
Finally, HaluAgent performs a label check to determine whether the hallucination label satisfies the definition of the hallucination type.
We manually design verification rules for each hallucination type.
If the generated data does not satisfy our proposed rules, we consider it as unqualified data.
If HaluAgent is uncertain about the correctness of the input, it filters out such cases to ensure the quality of the generated data.
Detailed rules are provided in Appendix \ref{sec:rules}.

\subsection{Prompt Optimization}
Despite careful design, existing studies have shown that directly prompting LLMs (even GPT-4) to generate QA data still results in suboptimal quality \citep{long-etal-2024-llms, anahv2}.
We believe this might stem from the incomprehensive of the initial prompt and its stylistic inconsistency with the LLM.
With meticulously designed requirements and a selection of examples for data generation, only $62.50\%$ of the data generated by Qwen-72B model with the initial few-shot prompting is qualified in a sampled subset (Table \ref{tab:optimization}).
The low validation rate brings two major problems:
1) \textbf{Reduced data volume}: Due to the high rejection rate of generated data, the actual number of qualified evaluation samples we obtain is reduced.
2) \textbf{Increase of resource consumption}: A large amount of computing resources are wasted on the generation of unqualified data.

Therefore, improving the validation rate of generated data is critical for automating the process of data construction.
We introduce prompt optimization to refine the prompt for data generation, making it more detailed and accurate, and better aligned with the LLM style.
Our proposed prompt optimization follows a multi-round iterative framework.
The optimization process samples a small number of knowledge documents to construct the training and validation datasets.
We use the verification module to output error feedback for unqualified data and introduce an optimization agent to help optimize prompts.
In each round of optimization, the optimization agent modifies the generation prompt according to the error feedback and generates a set of candidate optimized prompts.
We select the optimal prompt for the next round of optimization based on the validation rate of the generated data conditioned on the candidate prompts.

\subsection{Aggregation for Multiple QA Formats}
With the optimized prompt, we employ HaluAgent for large-scale data generation.
For each input knowledge document, we first generate the raw data needed for constructing QA pairs, including the question, the correct answer, three possible hallucinated responses, and the hallucination labels.
We construct questions for single-choice QA and true-or-false QA with the templates.
In single-choice QA, we randomly shuffle the order of the options to prevent potential bias.

\begin{table}[t]
\begin{center}
\resizebox{1.0\columnwidth}{!}{
\begin{tabular}{ccccccc}
\toprule
 & \bf Corr. & \bf Halu. & \bf Label & $\mathbf{\alpha}_\mathbf{corr}$ & $\mathbf{\alpha}_\mathbf{halu}$ & $\mathbf{\alpha}_\mathbf{label}$ \\
\midrule
W/o Ver & 96.33\% & 92.00\% & 76.00\% & 0.87 & 0.79 & 0.82 \\
W/Ver & 98.67\% & 99.00\% & 94.33\% & 0.80 & 0.89 & 0.71 \\
\bottomrule
\end{tabular}
}
\caption{Human annotation results on the verified QA data. \textbf{Corr.}, \textbf{Halu.} and \textbf{Label} denote the accuracy of correction check, hallucination check and label check given by human annotators respectively. We also provide $\mathbf{\alpha}_\mathbf{corr}$, $\mathbf{\alpha}_\mathbf{halu}$ and $\mathbf{\alpha}_\mathbf{label}$, which denote the Krippendorff's alpha \citep{krippendorff2004reliability} of human annotation.
}
\vspace{-7mm}
\label{tab:human}
\end{center}
\end{table}

\subsection{HaluAgent Results}
To assess the effectiveness of HaluAgent, we conduct experiments to illustrate the contribution of the verification module and evaluate the impact of prompt optimization.

\textbf{The verification module in HaluAgent effectively detects problematic data.}
To analyze the role of the verification module in HaluAgent, we compare the quality of the generated data with and without the verification module through human evaluation.
Human annotators are recruited to label the correctness of the generated correct answers, the hallucinated responses, and the hallucination labels given the extracted knowledge.
We randomly sample 300 generated correct answers, hallucinated responses, and hallucination labels from the verified dataset and the original dataset respectively for evaluation. 
We employ three human annotators for each sample.
The human annotators are provided with the designed rules to guide their annotations.
In cases of inconsistent annotations, we adopt a voting mechanism to select the majority label as the final human annotation result.
Detailed instructions for human evaluation are provided in Appendix \ref{sec:human_annotation}.
Table \ref{tab:human} provides the comparison of data correctness with and without the verification module.
The verification module in HaluAgent effectively identifies errors in the generated data, thereby improving the accuracy evaluated by human annotation.

\begin{table}[t]
\begin{center}
\resizebox{1.0\columnwidth}{!}{
\begin{tabular}{ccccc}
\toprule
\bf Iteration & \bf Corr.$\uparrow$ & \bf Halu.$\uparrow$ & \bf Label$\uparrow$ & \bf Overall$\uparrow$ \\
\midrule
0 & 97.50\% & 88.00\% & 71.00\% & 62.50\% \\
1 & 97.10\% & 95.65\% & 85.51\% & 81.16\%\\
2 & \bf 98.55\% & \bf 98.55\% & \bf 88.41\% & \bf 88.41\% \\
3 & 95.38\% & 92.31\% & 87.85\% & 86.15\% \\
\bottomrule
\end{tabular}
}
\caption{The validation rate of generated data on the validation set after each round of prompt optimization. Corr., Halu. and Label denote the pass rate of correction check, hallucination check and label check respectively. The overall column illustrates the pass rate of all three checks. 
}
\vspace{-9mm}
\label{tab:optimization}
\end{center}
\end{table}

\textbf{Prompt optimization helps improve data generation quality.}
As human evaluation demonstrates the effectiveness of the verification module, we subsequently adopt the verification module to evaluate and enhance the performance of the generation module.
We sample 60 and 20 knowledge documents from various domains as the training and validation inputs.
We set the maximum number of prompt optimization iterations to five.
If the validation rate on the validation set decreases during the optimization process, we terminate the prompt optimization early and select the optimal prompt on validation set as the final result.
As illustrated in Table \ref{tab:optimization}, we provide the performance of the generated data at different iteration stages during the prompt optimization process.
Through two rounds of iteration, the generation prompt achieves the best performance on the validation set, with an overall success rate of $88.41\%$, surpassing the initial result by $25.91\%$.

\begin{figure}[t]
    \centering
    \includegraphics[width=0.9\linewidth]{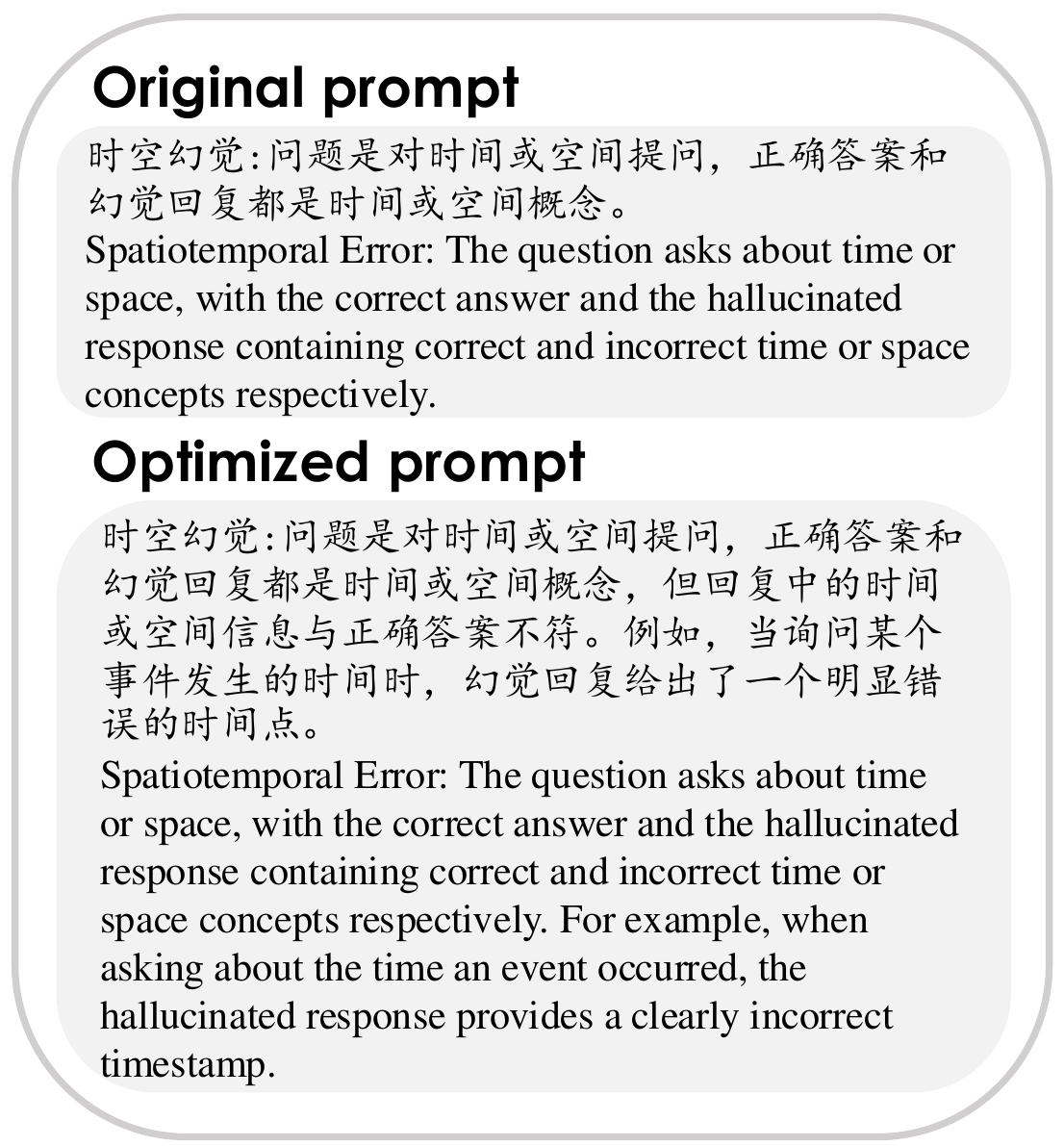}
    \caption{The prompt optimization case of spatiotemporal error definition in hallucination label generation.}
    \vspace{-5mm}
    \label{fig:case}
\end{figure}

\textbf{Optimized prompt enhances the clarity and details of the prompt.}
To investigate how HaluAgent modifies the prompt during the prompt optimization process, we manually examine prompt cases during optimization.
Figure \ref{fig:case} presents the original and optimized prompts for defining spatiotemporal error when generating hallucination labels.
Prompt optimization adds more details and requirements to the generation process, making the objectives more clear.
In addition, the optimization incorporates examples into the generated content to enhance the clarity of the explanations.
The complete results of prompt optimization are presented in Figure \ref{fig:opt_prompt} in the Appendix.

\section{C-FAITH: A Chinese Fine-Grained Question Answering Benchmark}
Using our HaluAgent method, detailed in Section \ref{sec:haluagent}, we introduce C-FAITH, a Chinese fine-grained QA hallucination benchmark.
By converting knowledge documents from various domains into question answering evaluation data, C-FAITH covers multiple domains.
This benchmark provides a scalable resource for Chinese hallucination evaluation with fine-grained error types.

\begin{table}[h]
\centering
\setlength{\tabcolsep}{5mm}
\resizebox{1.0\columnwidth}{!}{
\begin{tabular}{c|ccc}
\toprule
Overall & Gen. & Choice. & True/False. \\
\midrule
60,702 & 16,713 & 10,563 & 33,426 \\
\bottomrule
\end{tabular}
}
\caption{
Data Statistics of our proposed C-FAITH.
}
\vspace{-5mm}
\label{tab:stat}
\end{table}

\subsection{Dataset Overview}
We retrieve 1,399 knowledge documents from several Chinese databases including Wikipedia-CN\footnote{\href{https://zh.wikipedia.org/}{https://zh.wikipedia.org/}}, Baidu Baike\footnote{\href{https://baike.baidu.com/}{https://baike.baidu.com/}}, Sougou Baike\footnote{\href{https://baike.sogou.com/}{https://baike.sogou.com/}}.
These documents cover six major topics: Celebrities, Entertainment, Education, Astogeography, Biology, and Culture. 
We categorize the hallucination type of a question based on the primary hallucination category of the generated hallucinated responses. 
This hallucination type represents the specific type of hallucination the question is designed to induce\footnote{This hallucination type is the type that LLMs are most likely to generate and can also be referred to as the primary hallucination type.}. 
HaluAgent generates various types of questions from each knowledge document to elicit different hallucinations. 
Since hallucinated responses to the same question may be highly similar, we perform deduplication for single-choice QA. 
Only single-choice QA data containing distinct hallucinated options are retained. 
The statistics and distribution of the C-FAITH dataset are shown in Table \ref{tab:stat} and Figure \ref{fig:topic_stat}, including the distribution of topics and hallucination types for the questions in C-FAITH. 
Finally, we evaluate various LLMs using the proposed C-FAITH dataset.

\begin{table}[t]
\begin{center}
\resizebox{1.0\columnwidth}{!}{
\begin{tabular}{c|c|c|c|c}
\toprule
\bf Model & \bf Gen. $\downarrow$ & \bf Choice. $\downarrow$ & \bf True/False. $\downarrow$ & \bf Ova. $\downarrow$ \\
\midrule
GPT-4o & \underline{25.17} & \bf 10.40 & \bf 14.30 & \bf 16.53 \\
GPT-4 & 43.43 & 22.58 & 22.85 & 28.88 \\
GPT-3.5 & 45.69 & 42.60 & 38.70 & 45.19 \\
Doubao-pro-32k & 34.69 & 17.42 & 22.51 & 23.96 \\
moonshot-v1-8k & 44.42 & 25.48 & 26.93 & 31.99 \\
DeepSeek-V2.5 & 29.65 & 17.62 & 19.35 & 22.31 \\
DeepSeek-V3 & \bf 24.25 & \underline{11.60} & 16.70 & \underline{17.52} \\
ChatGLM-6B & 65.79 & 37.30 & 43.90 & 44.80 \\
GLM4-9B & 50.55 & 31.10 & 33.80 & 36.01 \\
LLaMA-3.1-8B & 67.75 & 41.60 & 33.15 & 45.62 \\
LLaMA-3.1-70B & 56.69 & 19.30 & 28.25 & 36.79 \\
LLaMA-3.3-70B & 44.88 & 19.20 & 23.20 & 28.77 \\
Yi-1.5-34B & 43.26 & 23.70 & 24.90 & 29.26 \\
Qwen-2.5-14B & 58.06 & 19.60 & 24.85 & 29.46 \\
Qwen-2.5-32B & 41.22 & 17.90 & 21.25 & 27.13 \\
Qwen-2.5-72B & 32.63 & 14.40 & \underline{16.40} & 20.82 \\
\bottomrule
\end{tabular}
}
\caption{The hallucination rate(\%) of the generated content from various LLMs. We calculate the hallucination rate of LLMs across the three formats of QA data and computed their average as the overall hallucination rate. We bold the best-performing model and underline the second-best-performing model.}
\vspace{-7mm}
\label{tab:general_results}
\end{center}
\end{table}

\begin{table*}[t]
\begin{center}
\resizebox{1.8\columnwidth}{!}{
\begin{tabular}{cccccccc}
\toprule
\bf Model & \bf Overall$\downarrow$ & \bf FactFab$\downarrow$  & \bf AttrErr$\downarrow$ & \bf EntErr$\downarrow$ & \bf RelErr$\downarrow$ & \bf SpaErr$\downarrow$ & \bf RefErr$\downarrow$ \\ 
\midrule
GPT-4o & \underline{25.17} & \bf 22.43 & \underline{13.01} & \underline{31.77} & \bf 15.15 & \bf 35.68 & \bf 10.00 \\
GPT-4 & 43.43 & 38.82 & 24.49 & 55.71 & 30.30 & 54.92 & 30.00 \\
GPT-3.5 & 54.31 & 52.96 & 33.61 & 67.95 & 30.30 & 65.36 & 30.00 \\
Doubao-pro-32k & 34.69 & 44.44 & 21.43 & 38.61 & 35.71 & 40.89 & 30.00 \\
moonshot-v1-8k & 44.42 & 45.20 & 25.04 & 53.71 & 36.36 & 57.55 & 30.00 \\
DeepSeek-V2.5 & 29.65 & 30.96 & 14.62 & 35.30 & \underline{18.18} & 41.97 & 23.33 \\
DeepSeek-V3 & \bf 24.25 & \underline{24.69} & \bf 10.98 & \bf 27.49 & \underline{18.18} & \underline{38.96} & 16.67 \\
ChatGLM3-6B & 65.79 & 67.49 & 50.50 & 72.80 & 42.42 & 77.98 & 50.00  \\
GLM4-9B & 50.55 & 53.25 & 29.03 & 59.89 & 36.36 & 65.80 & 36.67 \\
LLaMA-3.1-8B-Instruct & 67.75 & 65.33 & 50.17 & 78.43 & 42.42 & 80.05 & 50.00 \\
LLaMA-3.1-70B-Instruct & 56.69 & 55.73 & 41.95 & 65.38 & 60.61 & 64.77 & 36.67 \\
LLaMA-3.3-70B-Instruct & 44.88 & 41.80 & 27.35 & 56.04 & 30.30 & 55.70 & 26.67 \\
Yi-1.5-34B & 43.26 & 41.18 & 22.32 & 55.08 & 33.33 & 56.22 & 33.33 \\
Qwen-2.5-14B-Instruct & 41.94 & 41.80 & 19.63 & 51.99 & 39.39 & 59.07 & 20.00 \\
Qwen-2.5-32B-Instruct & 41.22 & 39.44 & 20.34 & 51.92 & 36.36 & 55.96 & 24.14 \\
Qwen-2.5-72B-Instruct & 32.63 & 33.75 & 16.11 & 39.42 & 27.27 & 45.85 & \underline{13.33} \\
\bottomrule
\end{tabular}
}
\caption{The hallucination rates(\%) of LLM responses across different hallucination types on generation QA.
}
\label{tab:halu_mode}
\end{center}
\end{table*}

\subsection{Experimental Settings}
\paragraph{Models} We evaluate 16 LLMs with C-FAITH, including GPT-4o, GPT-4 \citep{gpt4}, GPT-3.5\citep{gpt35}, Doubao-pro-32k, moonshot-v1-8k, DeepSeek-V2.5, DeepSeek-V3 \citep{deepseek-llm}, ChatGLM3-6B \citep{du-etal-2022-glm}, GLM4-9B \citep{glm4}, LLaMA-3.1-8B-Instruct, LLaMA-3.1-70B-Instruct, LLaMA-3.3-70B-Instruct \citep{llama3}, Yi-1.5-34B \citep{ai2024yiopenfoundationmodels}, Qwen-2.5-14B-Instruct, Qwen-2.5-32B-Instruct and Qwen-2.5-72B-Instruct \citep{qwen2.5}.
We adopt the same decoding settings (temperature=1, top\_p=0.7) for all LLMs.

\paragraph{Metrics} 
The evaluation is conducted in the zero-shot setting.
We calculate the hallucination rate of LLM generations as the evaluation metric.
For generative QA, to evaluate whether the LLM output contains hallucination, we employ GPT-4o to determine if the LLM output conflicts with the correct response.
If a conflict is identified, the LLM output is considered to be hallucinated.
For single-choice QA and true-or-false QA, we determine the correctness of the LLM output with the correct answer by directly comparing it with the correct answer.
We calculate the average hallucination rate across the three QA formats as the overall hallucination rate for each LLM.

\begin{figure*}[t]
    \centering
    \includegraphics[width=0.8\linewidth]{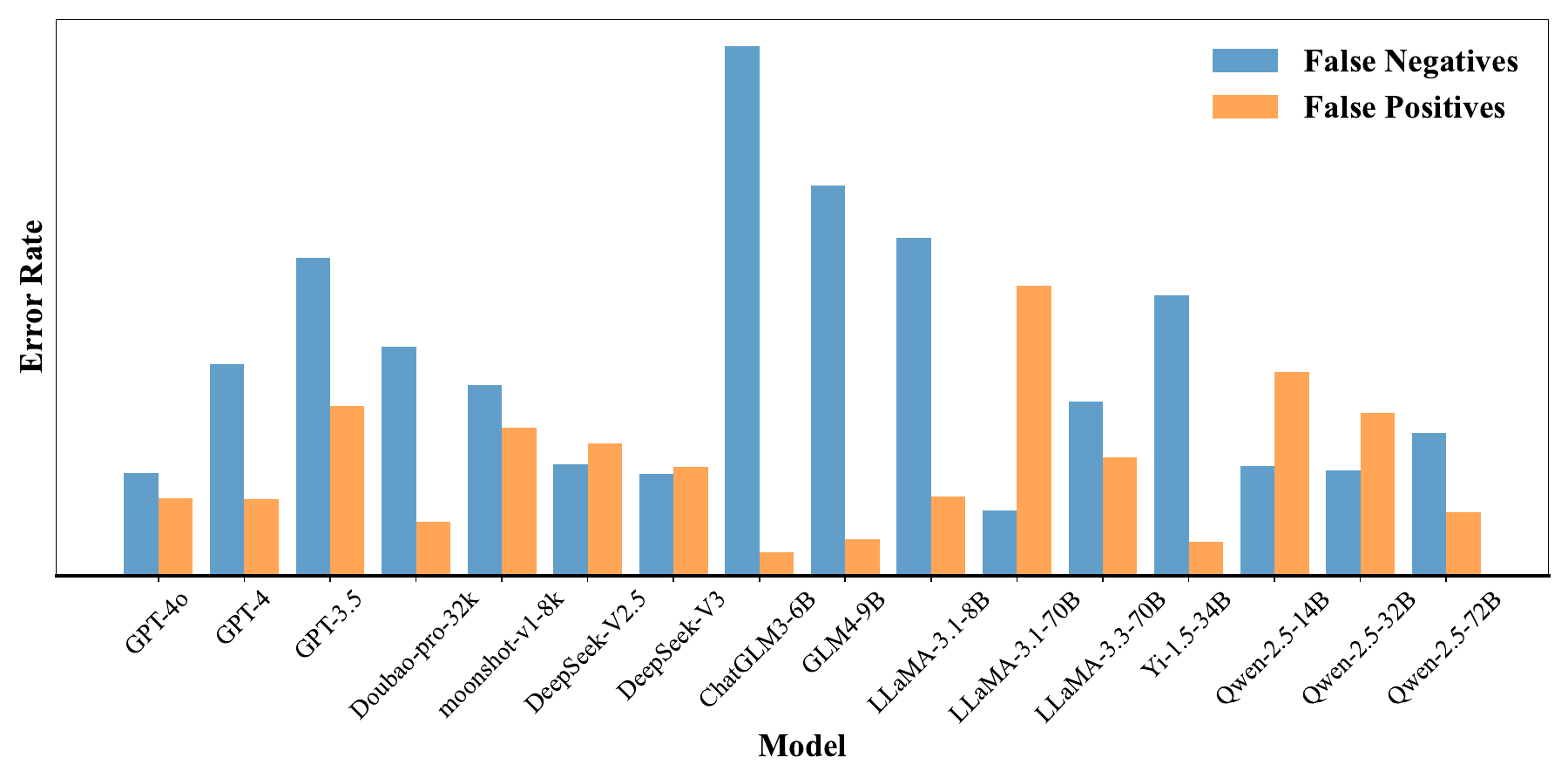}
    \caption{The false negative and false positive error rates of LLMs when facing true-or-false questions.}
    \vspace{-3mm}
    \label{fig:error}
\end{figure*}

\subsection{Results and Analysis}
Table \ref{tab:general_results} presents the experimental results for 16 different LLMs. 
In general, GPT-4o shows the lowest overall hallucination rate, followed by DeepSeek-V3 and Qwen-2.5-72B-Instruct. 
Deepseek-v3 performs well in the the generative QA task achieving a lower hallucination rate than GPT-4o.
Among the various QA formats, LLMs are more prone to generating hallucinations in generative QA tasks, leading to lower response accuracy. 
In most cases, the models exhibit similar relative performance across the three QA formats. However, some LLMs, such as LLaMA-3.1-70B, show low accuracy in generative QA but perform well in single-choice QA, even achieving higher accuracy than GPT-4.

In addition, we observe the following insights:

\paragraph{Larger LLMs generally exhibit lower overall hallucination rate.}
As expected, scaling up model sizes typically leads to low hallucination rates.
Under the same model architecture, such as in the Qwen-2.5 series and LLaMA3 models, the hallucination rate decreases as the model's parameter size increases across all three QA formats.

\paragraph{A significant difference exists in hallucination rate across different hallucination types.}
As C-FAITH provides the hallucination label for each question, we analyze the impact of hallucination label on the hallucination rate of LLM response.
We focus on experimental results of generative QA data.
Table \ref{tab:halu_mode} summarizes the hallucination rate of LLM responses conditioned on different hallucination labels.
LLMs have their own strengths when responding to questions of different hallucination labels.
For example, the Qwen series achieve low hallucination rate when addressing quesitions related to attribute errors, while performimg poorly when handling spatiotemporal-related questions.

For the vast majority of LLMs, the probability of hallucinated generation is high when faced with questions meant to induce entity errors and spatiotemporal errors.
On one hand, this is because the answers to these types of questions are deterministic and unique.
Therefore, any discrepancy between the LLM generation and the facts results in hallucination.
On the other hand, LLMs are generally prone to entity confusion and erroneous memory of time and location information.
Through fine-grained hallucination classification, we help identify the main category of hallucination in LLMs, providing guidance for targeted hallucination mitigation strategies.

\paragraph{Most LLMs are more likely to be fooled by the input hallucinated content.}
For true-or-false QA, we categorize LLM errors into two types: false negative (where the model incorrectly classifies a hallucinated response as correct) and false positive (where the model incorrectly classifies a correct response as hallucinated).
In Figure \ref{fig:error}, we present a comparison of false negative and false positive error rates across different LLMs.
Except for a few LLMs such as LLaMA-3.1-70B, most models exhibit a higher probability of generating false negatives than false positives. 
Models such as ChatGLM-6B and LLaMA3.1-8B exhibit a significantly higher false negative error rate compared to the false positive error rate.
Even GPT-4 exhibits a clear bias to generate false negative errors.
This phenomenon suggests that most LLMs tend to accept the input content as valid, even when the input itself contains hallucinations.

\section{Conclusion}
We address the limitations in automated fine-grained question answering benchmarks for hallucination evaluation by introducing HaluAgent, a multi-agent system that automatically constructs hallucination benchmarks.
HaluAgent effectively creates questions, correct answers, hallucinated responses and hallucination labels according to the input knowledge document.
Building on this, C-FAITH provides a benchmark of 60,702 QA data in total, providing a new fine-grained scalable benchmark for LLM hallucination evaluation.

\section*{Limitations}
Currently, the C-FAITH evaluation dataset encompasses six types of hallucinations. We plan to refine and correct any misclassifications to ensure a more comprehensive evaluation of hallucinations. In addition, our dataset primarily focuses on six general knowledge domains. We aim to expand the evaluation to include other fields, such as healthcare and finance.
\bibliography{custom}

\appendix

\section{Manual rules in the verification module}
\label{sec:rules}
We provide the manual rules used in the verification module in this section.

\section{Human Evaluation}
\label{sec:human_annotation}
We provide our human evaluation guideline furnished to participants for manually evaluating the correctness of the generated correct answer, hallucinated response and hallucination label.
We recruited three Chinese college students for annotation.
\begin{tcolorbox}[colback=black!5!white,colframe=black!75!white,title=Human Evaluation Instructions]
\begin{CJK}{UTF8}{gbsn} 
\small 
感谢参与本次标注任务！我们目前正在进行一个幻觉研究项目，需要评价生成幻觉评测数据的正确性。我们生成的数据包含问题、正确答案、背景知识、幻觉答复和幻觉标签。我们需要您评价正确答案正确性（要求：正确答案是对问题的正确回答，背景知识可以视为正确的事实）、幻觉回复的幻觉性（要求：幻觉答案中包含与背景知识不一致的幻觉的表述）、幻觉标签正确性（要求：幻觉标签与给出的幻觉标签的定义相符合）。具体的幻觉标签的定义如下：

虚构事实：[幻觉回复]必须包含不存在的概念或无法被现有资料证实的伪事实。[幻觉回复]中包含的内容在现实世界中没有实际依据，或者是错误地捏造出来的。例如，完全虚构的事件、人物、地点或其他无法查证的事实。如果[问题]中明确要求描述事物的某些属性，如功能、特征、组成等时，一律归为属性错误而不符合虚构事实的要求。

属性错误：[问题]中明确要求描述事物的某些属性，如功能、特征、组成等，而[幻觉回复]中包含对这些实体属性的错误表述。

实体错误：[问题]应针对具体实体进行提问。实体指的是特定的、具有实际存在或明确身份的事物，例如人名、地名、作品名、事件名、概念等。[幻觉回复]包含与[正确答案]或已知世界知识相矛盾的错误实体，例如人名、事件名、书籍、电影等。实体错误专注于对已知实体的错误描述。实体通常是现实世界中已经明确存在的东西，所以这个错误不涉及虚构内容，而是对已经存在的事物进行错误的归类或描述。

时空幻觉：[幻觉回复]必须涉及时间、空间或特定时期的描述。时空幻觉仅限于时间或地点描述错误，不包括对事件本身的内容或属性的误解。如果对事件本身内容表述有误，归类于虚构事实或属性幻觉。

关系错误：[幻觉回复]必须包含对不同实体之间的比较，而不是单独对每个实体的描述。[幻觉答案]与[正确答案]在描述不同实体之间关系时，必须存在不一致。

虚假引用：[幻觉回复]中必须包含虚假的引用或链接。

\end{CJK}
\end{tcolorbox}

\onecolumn
\begin{tcolorbox}[colback=black!5!white,colframe=black!75!white,title=Manual rules for verification]
\begin{CJK}{UTF8}{gbsn} 
\small 
\textbf{正确答案检测：}

假设你是一个中文问答正确性检验者。请根据给定的背景知识，判断所给答案是否是问题的正确回复并给出理由。具体对问答的要求如下：

1. [问题]必须能够在[背景知识]中找到正确答案。

2. [正确答案]必须是对问题的正确回复。

3. [正确答案]必须与背景知识相符，背景知识可以视为正确的事实。

4. [正确答案]自身必须符合事实和逻辑，足够合理。

<例子>

\textbf{幻觉回复检测：}

 假设你是一个中文问答错误性检验者。请根据给定的背景知识，判断所给幻觉答案是否是问题的错误回复并给出理由。具体对问答的要求如下：
 
1. [幻觉答案]必须是对问题的错误回复。

2. [幻觉答案]必须与背景知识相违背，或与背景知识中的逻辑和信息不符。

3. [幻觉答案]自身必须符合事实和逻辑，足够合理。

<例子>

\textbf{幻觉标签检测：}
\textbf{虚构事实}：假设你是一个幻觉类型检测员。请根据给定的要求判断是否存在虚构事实，并提供理由。具体要求如下：

1. [幻觉答案]必须包含不存在的概念或无法被现有资料证实的伪事实。[幻觉答案]中包含的内容要么是完全不存在的概念，要么是无法通过现有可靠资料证实的伪事实。这些内容在现实世界中没有实际依据，或者是错误地捏造出来的。例如，完全虚构的事件、人物、地点或其他无法查证的事实。

2. [幻觉答案]与[正确答案]在事实或概念的表述上存在不一致，必须包含完全虚构的内容，与现实完全不符。[幻觉答案]中的内容应与[正确答案]在事实或概念上的表述完全不一致。这个不一致不是简单的误解或错误，而是包含了完全虚构的部分，和现实世界或公认的事实有根本的区别。例如，错误地描述一个不存在的历史事件，或提供一个没有任何依据的虚构数据。

3. [问题]中明确要求描述事物的某些属性，如功能、特征、组成等时，一律归为属性错误而不符合虚构事实的要求。

\textbf{属性错误}：假设你是一个幻觉类型检测员。请根据给定的要求判断是否存在属性错误，并提供理由。具体要求如下：

1. [问题]明确要求描述事物的某些属性，如功能、特征、组成等。[问题]本身会清楚地要求对某个具体实体或事物的属性进行描述。例如，问题可能询问某个物体的功能、组成、外观特征、用途、资格等。[问题]并不要求实体的整体或身份描述，而是聚焦于对该事物的某些具体属性的说明。

2. [幻觉答案]中包含对现实世界存在的物体进行错误的属性描述，通常是误导性的或完全不符合该事物的实际特性。[幻觉答案]将现实世界中的某个物体或事物的属性描述错误。这些错误的属性描述通常表现为错误的功能描述、错误的外观特征描述、错误的组成描述等。

3. [幻觉答案]与[正确答案]对实体属性的描述存在不一致。这种差异不仅仅是表述上的不一致，而是指对该事物核心属性的描述发生了根本错误或误解。例如，描述某个物体的特征时，幻觉答案的描述与正确答案显著不符，导致用户得到一个错误的认知。    

\textbf{实体错误}：假设你是一个幻觉类型检测员。请根据给定的要求判断是否存在实体错误，并提供理由。具体要求如下：

1. [问题]应针对具体实体进行提问。实体指的是特定的、具有实际存在或明确身份的事物，例如人名、地名、作品名、事件名、概念等。[问题]应聚焦于这些实际存在的具体对象，而不是抽象概念。

2. [幻觉答案]包含与[正确答案]或已知世界知识相矛盾的错误实体，例如人名、事件名、书籍、电影等。实体错误专注于对已知实体的错误描述。实体通常是现实世界中已经明确存在的东西，所以这个错误不涉及虚构内容，而是对已经存在的事物进行错误的归类或描述。例如，如果问题问某部电影的导演，而幻觉答案给出了错误的导演名字，尽管电影确实存在，错误的名字就构成了实体错误。

\textbf{时空幻觉}：假设你是一个幻觉类型检测员。请根据给定的要求判断是否存在时空幻觉，并提供理由。具体要求如下：

1. [幻觉答案]必须涉及时间、空间或特定时期的描述。[幻觉答案]需要明确描述与时间（如年份、日期、时刻等）或空间（如地点、地理位置等）相关的内容。这种描述通常与事件发生的时刻、地点、持续时间等信息密切相关。比如提到某事件发生的具体年份、发生地点或涉及的时间范围。

2. [幻觉答案]与[正确答案]在时间或空间的描述上存在不一致。[幻觉答案]中的时间或空间信息必须与正确答案的实际时间或空间信息不符。这种不一致通常表现为错误的年份、日期、时刻，或是错误的地点描述。

3. 时空幻觉仅限于时间或地点描述错误，不包括对事件本身的内容或属性的误解。[幻觉答案]中对事件或属性描述的具体内容必须正确，只有时间或空间部分出错。

\textbf{关系错误}：假设你是一个幻觉类型检测员。请根据给定的要求判断是否存在关系错误，并提供理由。具体要求如下：

1. [幻觉答案]必须包含对不同实体之间的比较，而不是单独对每个实体的描述。

2. [幻觉答案]与[正确答案]在描述不同实体之间关系时，必须存在不一致。

\textbf{虚假引用}：假设你是一个幻觉类型检测员。请根据给定的要求判断是否存在虚假引用，并提供理由。具体要求如下：

1. [幻觉答案]中必须包含引用或链接。

2. [幻觉答案]中的引用或链接必须是虚构的，实际上这些引用和链接并不存在，或与[正确答案]中给出的引用和链接不一致。

\end{CJK}
\end{tcolorbox}

\begin{table*}
\centering
\begin{tabular}{p{3cm}p{6cm}p{6cm}}
    \toprule
    \small
    \textbf{Hallucination type} & 
    \small
    \textbf{Question} &
    \small
    \textbf{Hallucinated Response} \\
    \midrule
    \small
    \textbf{FactFab} & 
    \small
    Tell me about the historical origins of unicorns. &
    \small
    Unicorns were documented to have roamed the plains of Atlantis around 10,000 BC, where they were considered sacred creatures and were often associated with royalty. \\
    \hline
    \small
    \textbf{AttrErr} & 
    \small
    Please introduce the function of a pen. &
    \small
    Pens are primarily used for painting. They are usually made of metal, and do not use ink. \\
    \hline
    \small
    \textbf{EntErr} & 
    \small
    Which film won the Palme d'Or at the 75th Cannes Film Festival? &
    \small
    The Palme d'Or at the 75th Cannes Film Festival was awarded to the French film The Noon Star. \\
    \hline
    \small
    \textbf{RelErr} & 
    \small
    Who was born first, Aaron Gillespie or Nathan Leone? &
    \small
    Aaron Gillespie was born before Nathan Leone. \\
    \hline
    \small
    \textbf{SpaErr} & 
    \small
    What is Rembrandt Halmensson van Rijn's date of birth? &
    \small
    June 14, 1605. \\
    \hline
    \small
    \textbf{RefErr} & 
    \small
    Give me five of the most influential research articles on large model hallucinations. &
    \small
    Towards a Rigorous Science of Neural Language Models" by LeCun et al. (which actually does not exist) \\
    \bottomrule
    \end{tabular}
    \caption{Examples of specific subtypes of hallucinations.}
    \label{tab:halu_type}
\end{table*}

\begin{figure*}[t]
    \centering
    \subfigure[Topic distribution of C-FAITH]{
    \includegraphics[scale=0.38]{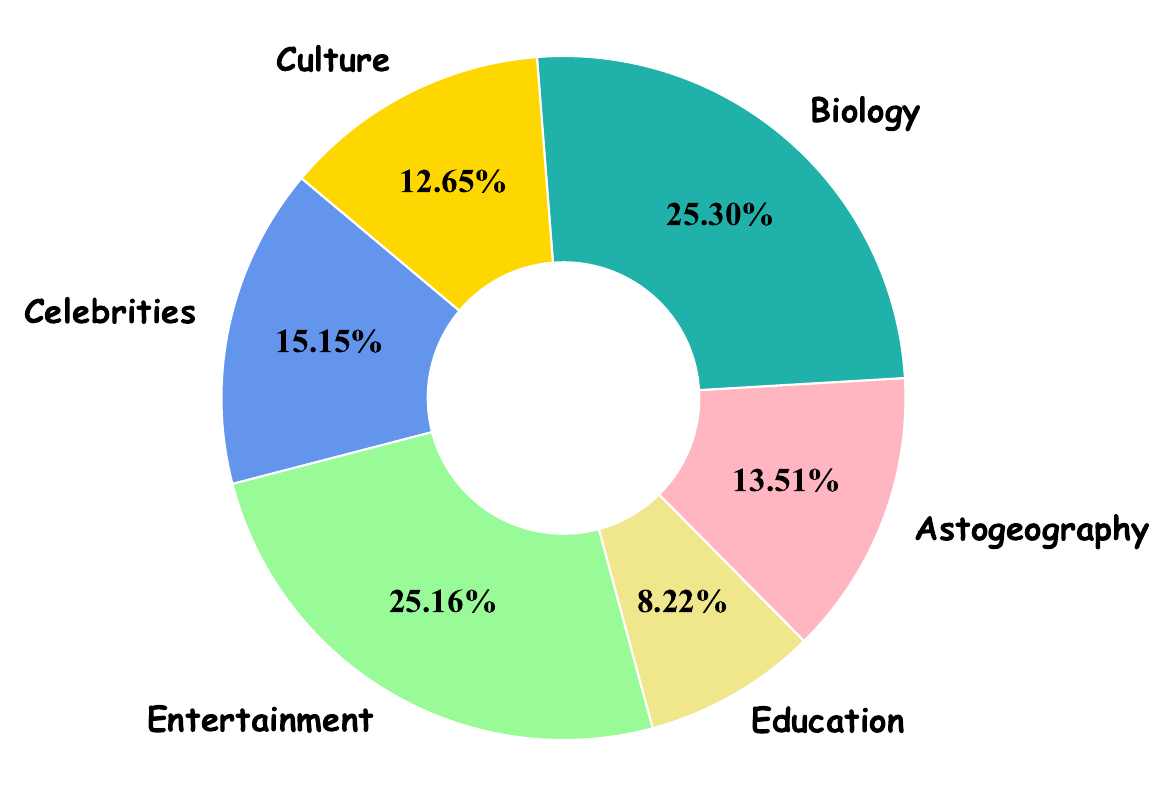}
    }
    \subfigure[Hallucination type distribution of C-FAITH]{
    \includegraphics[scale=0.38]{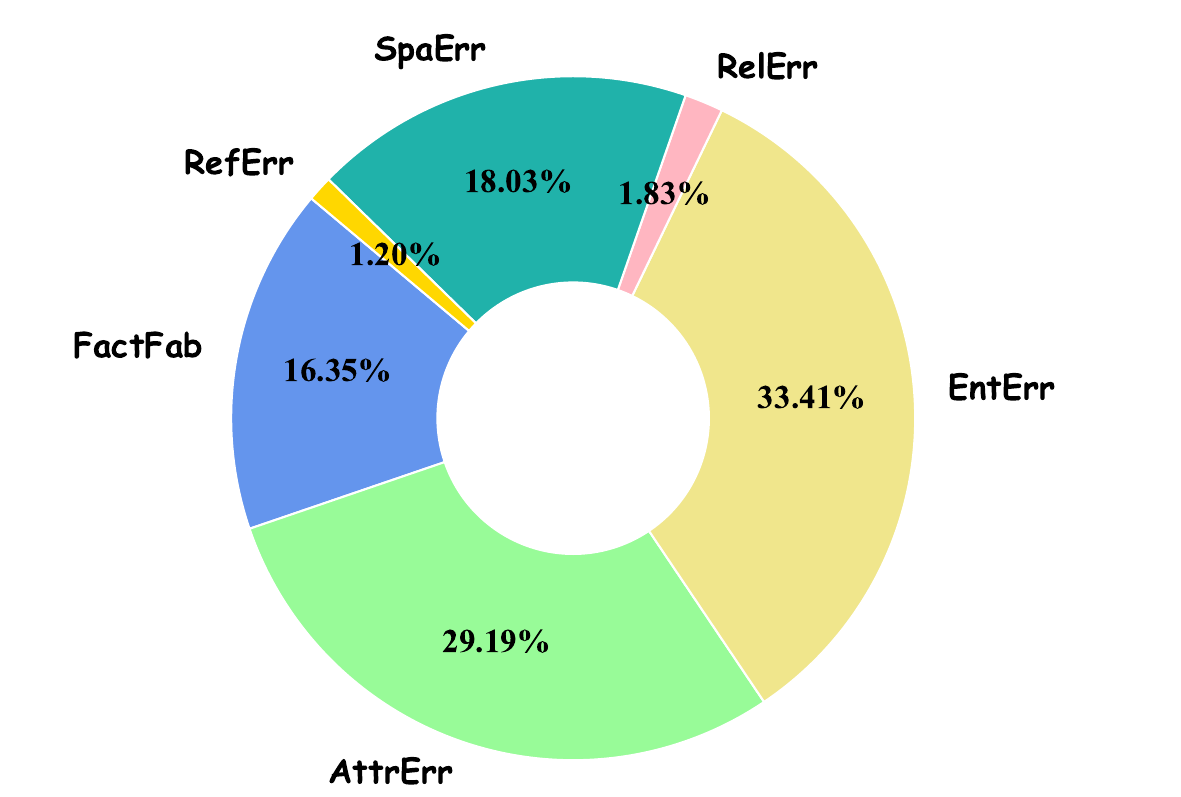}
    } 
    \caption{General Overview of C-FAITH dataset, containing topic and hallucination type distribution of the dataset.}
    \label{fig:topic_stat}
\end{figure*}

\begin{figure*}[t]
\centering
    \includegraphics[width=1.0\linewidth]{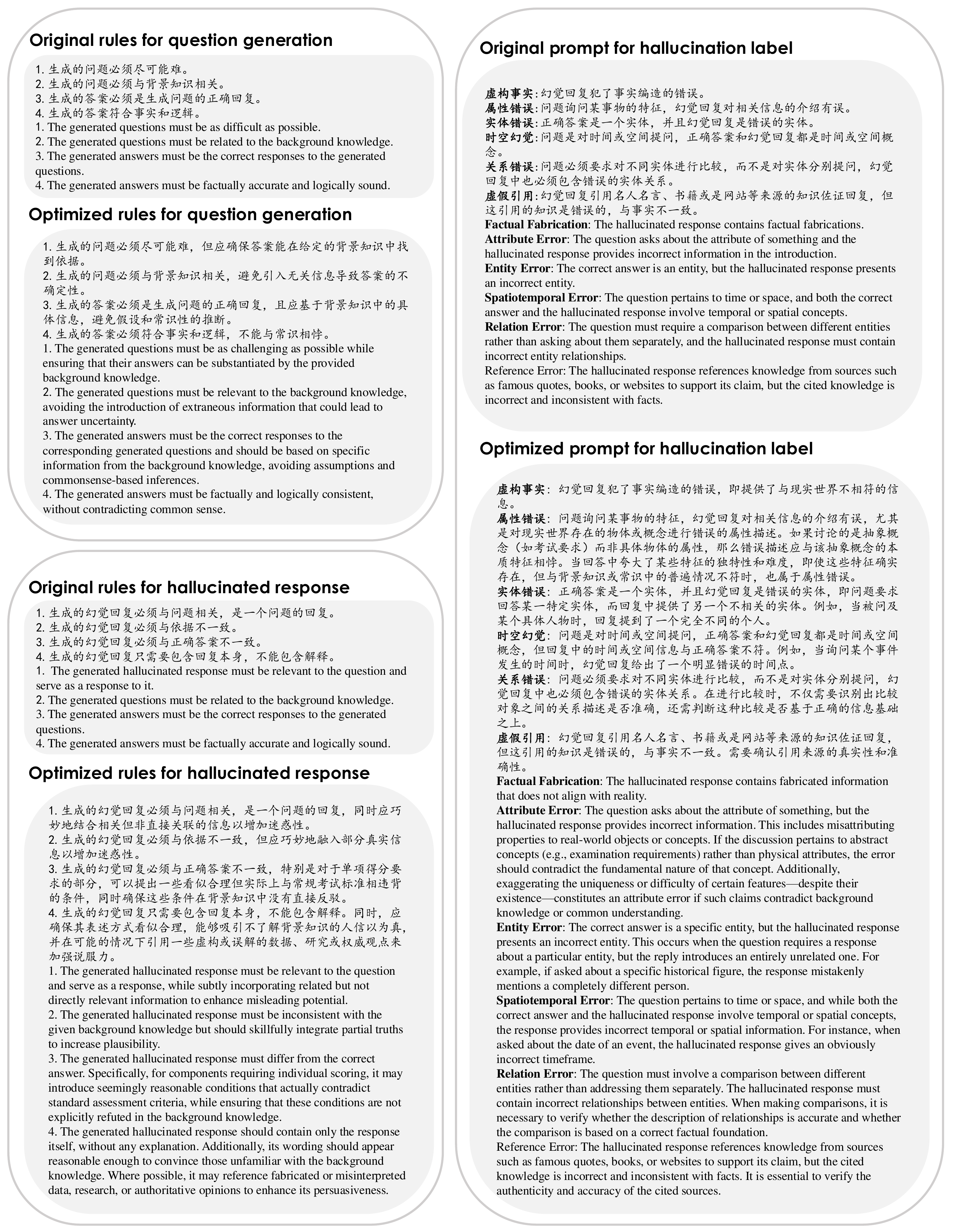}
    \caption{Original prompt and optimized prompt for data generation.}
\label{fig:opt_prompt}
\end{figure*}

\end{document}